\documentclass[twocolumn]{article}
\usepackage[dvipdfmx]{graphicx}
\usepackage{arxiv}
\usepackage{authblk}

\usepackage{multirow}%
\usepackage{amsmath,amssymb,amsfonts,amsthm}
\usepackage{mathrsfs}%
\usepackage[utf8]{inputenc}
\usepackage{hyperref}
\usepackage{enumitem}

\usepackage{xcolor}%
\usepackage{textcomp}%
\usepackage{manyfoot}%
\usepackage{booktabs}%
\usepackage{algorithm}%
\usepackage{algorithmicx}%
\usepackage{algpseudocode}%
\usepackage{listings}%

\usepackage{multirow}
\usepackage{subfig}
\usepackage{array}
\usepackage{booktabs}
\usepackage{threeparttable}
\usepackage{colortbl}

\theoremstyle{thmstyleone}%
\theoremstyle{thmstyletwo}%

\theoremstyle{thmstylethree}%
\newcommand{\ccol}[2]{ \multicolumn{#1}{c}{#2}}
\newcommand{\graycell}{\cellcolor[gray]{0.8}}

\newcolumntype{P}[1]{>{\centering\arraybackslash}m{#1}}

\newcommand{\colwid}{1.1cm}

\begin{document}

\newcommand{\myPaperShortTitle}{Stable LLM Ensemble}
\newcommand{\myPaperTitle}{Stable LLM Ensemble: Interaction between\\Example Representativeness and Diversity}
\title{\myPaperTitle}
\date{}

%\author{Junichiro Niimi}

\renewcommand\Authfont{\bfseries}
\setlength{\affilsep}{0em}
% box is needed for correct spacing with authblk
\newbox{\orcid}\sbox{\orcid}{\includegraphics[scale=0.06]{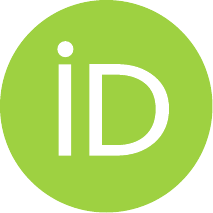}} 
\author[1,2]{%
	\href{https://orcid.org/0000-0002-4618-6272}{\usebox{\orcid}\hspace{1mm}
	Junichiro Niimi\thanks{\texttt{jniimi@meijo-u.ac.jp}}
	}}
\affil[1]{Meijo University}
\affil[2]{RIKEN AIP}

\renewcommand{\shorttitle}{\myPaperShortTitle}
\newcommand{\FOne}{macro\text{-}F_1}
\newcommand{\RMSE}{R\hspace{-0.05em}M\hspace{-0.1em}S\hspace{-0.07em}E}

%\maketitle

\twocolumn[
	\begin{@twocolumnfalse}
		\maketitle
\vspace{-3em}
\begin{abstract}
Large language models (LLMs) have achieved remarkable results in wide range of domains. However, the accuracy and robustness of one-shot LLM predictions remain highly sensitive to the examples and the diversity among ensemble members. This study systematically investigates the effects of example representativeness (one-shot strategy) and output diversity (sampling temperature) on LLM ensemble performance. Two one-shot strategies are compared: centroid-based representative examples (proposed) and randomly sampled examples (baseline) and sampling temperature also is varied. The proposed approach with higher temperature setting significantly outperforms random selection by $+7.6\,\%$ (macro-F1) and $-10.5\,\%$ (RMSE). Furthermore, the proposed model exceeds 5-shot prompting by $+21.1\,\%$ (macro-F1) and $-24.0\,\%$ (RMSE). Our findings demonstrate that combining representative example selection with increased temperature provides the appropriate level of diversity to the ensemble. This work highlights the practical importance of both example selection and controlled diversity in designing effective one-shot LLM ensembles.
\end{abstract}
\vspace{0.5em}
\vspace{2em}
	\end{@twocolumnfalse}
]

\renewcommand\thefootnote{*}
\setcounter{footnote}{0}
\section{Introduction}
\subsection{Background}
Large Language Models (LLMs) have achieved remarkable performance across a wide range of natural language processing (NLP) tasks. In particular, there is a rapid expansion of LLM applications into diverse domains, including marketing \cite{llm_sentiment_review,llm_marketing_review}, finance \cite{llm_finance_review,llm_finance_review2}, and education \cite{llm_education_review}. These models have enabled novel approaches to various tasks such as sentiment analysis, machine translation, retrieval-augmented generation (RAG), and text summarization. As a result, LLMs are fundamentally transforming the way data-driven insights are generated and utilized in each field.

Despite their impressive abilities, the output of LLMs remains highly sensitive to model configuration, such as one-/few-shot learning \cite{llm_sensitive_to_example}, prompt templates \cite{llm_prompt_template},  and hyperparameters \cite{llm_calibration,llm_temperature}. The performance and consistency of LLM-generated predictions often depend on subtle differences in example selection and prompt construction. However, there is currently no established method for optimal example selection, and many existing studies still rely on random sampling strategies.

In addition, ensemble learning \cite{ensemble} using LLMs has been actively conducted \cite{llm_hetero2,niimi_nldb,llm_hetero}. Ensemble methods utilize the inherent randomness in machine learning including LLMs to achieve higher accuracy \cite{llm_hetero} and computational efficiency \cite{niimi_nldb}, compared to single inference. However, since the model configuration greatly affects the individual inference of LLMs, it is also important in ensemble.

\subsection{Research Gaps}
From the research background, the variability of the multiple inferences due to prompts, examples, and hyperparameters remains a major challenge for the practical application of sentiment analysis, particularly in terms of reproducibility and robustness. We derive five research questions (RQs 1--5) and approaches as follows:

\paragraph{RQ1 Effect of temperature on ensemble effectiveness:} Compare to the lower temperature, do the high temperature setting contribute to the model performance?\\
{\it Approach:} We systematically vary the sampling temperature parameter in the LLM, evaluating ensemble accuracy, such as F1 and root mean square error (RMSE) at low (temperature=0.8) and high (temperature=1.5) settings for each one-shot approach and examine the change in accuracy.

\paragraph{RQ2 Effect of example representativeness on ensemble performance:} Compared to the random example, does the representative example contribute to the model performance?\\
{\it Approach:} We set up two one-shot strategies: Centroid-based Representative Examples (CREs, proposed) and Randomly-Sampled Examples (RSEs, baseline) and compare accuracy between the approaches.

\paragraph{RQ3 Optimal combination of representativeness and diversity:} Which combination of temperature and example selection achieves the best overall performance?\\
{\it Approach:} We overlook the all combinations of model configuration over the one-shot approaches and temperature settings to identify which configuration achieves the highest model performance.

\paragraph{RQ4 Comparison with 5-shot model:} Which is more effective, the ensemble model or the 5-shot model?\\
{\it Approach:} From the results of the optimal configuration, we additionally compare the performance between ensemble and 5-shot models.

\paragraph{RQ5 Effect of consistency:} In the best ensemble model, how does the self-consistency among base model predictions relate to the improvement achieved by ensemble aggregation?\\
{\it Approach:} We conduct an in-depth analysis of the best-performing model to investigate how ensemble performance relates to the self-consistency among base model predictions. Specifically, we examine whether improvements achieved by ensemble aggregation are associated with the degree of agreement or disagreement among the individual base models.

To address these issues, in this study, we propose the method to select representative examples from the dataset and examine the effect to utilize such sample on accuracy of LLM-based ensemble. In addition, we also vary the sampling temperature parameter to examine the effect of the creativity of the inference. We figure out the optimal model setting for the LLM-ensemble methods in terms of prediction accuracy and robustness and further compare with 5-shot settings.

\section{Related Works}
\subsection{Sentiment Analysis}
Sentiment analysis \cite{sentiment_classic}, which aims to identify and classify subjective information from text, has long been a major topic in natural language processing (NLP) and machine learning.
Early approaches were based on traditional machine learning methods such as logistic regression, support vector machines (SVM) \cite{svm}, and Naive Bayes classifiers \cite{nb}, often combined with hand-crafted features or shallow linguistic representations \cite{sentiment_review_2}.

With the advent of deep learning, DNN-based methods, such as convolutional neural networks (CNNs) \cite{cnn} and recurrent neural networks (RNNs) \cite{rnn}, have become the standard baselines, enabling more accurate sentiment prediction by capturing complex semantic patterns in text \cite{sentiment_review_2}.
These methods rely on large annotated datasets and powerful architectures to achieve state-of-the-art results.

More recently, the emergence of LLMs such as GPT and BERT has brought about a new paradigm for sentiment analysis \cite{llm_sentiment_review}.
LLMs are capable of both zero-shot and few-shot learning, making it possible to perform sentiment analysis without task-specific fine-tuning.
A number of studies have explored various aspects of LLM-based sentiment analysis, including the design of effective prompts \cite{gpt35_sentiment}, comparisons with human annotators \cite{gptVsMTurk}, and the impact of one-shot example selection on model performance \cite{llm_sensitive_to_example,automatic_example}.
There is growing evidence that LLM-based approaches can outperform conventional ML and DNN-based methods on diverse sentiment analysis tasks \cite{llm_sensitive_to_example,niimi_nldb,gpt35_sentiment}.

\subsection{Ensemble in LLMs}
Ensemble learning is referred to as a combination of multiple classifiers to construct a robust predictor \cite{ensemble,ensemble2}. To consolidate multiple predictions, several techniques are adopted: voting \cite{averaging} (e.g., majority voting or weighted voting, where predictions are aggregated based on the predefined weights), bagging \cite{bagging,randomforest} (bootstrap aggregating, where weak learners are trained on different subsamples of a training set), and boosting \cite{adaboost,gradientBoosting} (e.g., AdaBoost and Gradient Boosting, where weak learners are iteratively improved based on previously misclassified samples). These methods utilizes the diversity among the individual classifiers, as they are particularly effective for unstable learning algorithms \cite{ensemble}. 

In regard to LLM inference, prior studies have implemented the ensemble approaches \cite{llm_ensemble_survey,llm_ensemble_productattribute,llm_hetero2,niimi_nldb,llm_hetero}. For example, one financial study \cite{llm_hetero} employed a stacked LLM architecture which lower LLMs (base models) were assigned the multiple characteristics to ensure the diversity and upper LLM consolidated the decisions. Another marketing study \cite{niimi_nldb} diversified the base models by using different random seed values and consolidated the decisions with median. 
While the former study \cite{llm_hetero} is notable for promoting model diversity by assigning different areas of expertise through prompt engineering, this strategy requires manual design and careful consideration of which expertise to assign to each model. Such manual intervention introduces additional overhead and the potential for subjective bias in model configuration. In addition, the aggregation mechanism it employs further exacerbates the inherent black-box nature of LLM-based decision making.

In contrast, the latter study \cite{niimi_nldb} leverages median aggregation to minimize the influence of outliers in ordinal sentiment classification, but the diversity among base models is limited, as it only varies random seed values rather than introducing substantive differences between models.

\subsection{Consistency and Reliability in LLMs}
Moreover, LLMs have a temperature parameter which can control the randomness of the output. Some studies \cite{llm_selfconsistency,annotation_gpt4} highlighted the importance of self-consistency of the inferences. In particular, the consistency of the multiple inferences in higher temperature setting decreased, resulting in more diverse predictions for the same input \cite{annotation_gpt4}. While their work focuses on automated annotation and uses “consistency score” to identify edge cases, their findings suggest that sampling temperature can be used as a mechanism to systematically control output diversity in LLMs. 

Recent work has also addressed the calibration and uncertainty quantification of LLM predictions, proposing methods to assess and improve model reliability in various NLP tasks \cite{llm_uncertainty1,llm_calibration,llm_uncertainty2}. These approaches enable the use of consistency-based metrics for dynamic inference, error flagging, and confidence-aware downstream applications.

\section{Proposed Model}
In this study, we first choose the appropriate examples based on the embedding clusters to sufficiently ensure the diversity. Second, we conduct model inference with varying sampling temperature and consolidate multiple decisions with median. The overview of our configuration is shown in Table \ref{tab:setting}.

\begin{table}[htb]
   \centering
      \caption{Model Setting}\label{tab:setting}
\begin{tabular}{ll}
\toprule
\ccol{1}{Parameters} & \ccol{1}{Value} \\
\midrule
Random seeds & $\{1, 2, 3, 4, 5\}$ \\
Sampling temperature & $\{0.8, 1.5\}$ \\
top\_p & 0.9\\
max\_new\_tokens & 1\\
\bottomrule
\end{tabular}
\end{table} 

\subsection{Base Models}
For ensemble inference, we constructed five base models ($M_1$--$M_5$), each defined by a unique combination of one-shot example and random seed. Specifically, we first selected five distinct one-shot examples (Example 1--5) from the data pool, according to the specified strategy (either CREs or RSEs). For each base model $M_i$, where $i \in \{1, 2, 3, 4, 5\}$, we assigned Example $i$ and set the random seed $i$ (e.g., $M_1$ uses Example 1 and seed 1). In addition, to stochastically vary the responses, we use nucleus sampling \cite{nucleus_sampling}.
This design ensures that each base model is exposed to a different example and stochastic generation path, thereby promoting diversity among ensemble members while maintaining reproducibility.

First, regarding the base models for ensemble, this study employs Meta AI's Llama family \cite{llama3} (Llama-3.1-8B-Instruct)\footnote{https://huggingface.co/meta-llama/Llama-3.1-8B-Instruct} for the pretrained model. Llama is trained with large corpus with reinforcement learning from human feedback (RLHF) \cite{rlhf} and has been widely adopted in the sentiment analysis. In accordance with prior studies \cite{llm_hetero,niimi_nldb}, we do not perform fine-tuning. 

The prompt is shown in Fig. \ref{fig:prompt}, which the review and rating selected as an example is imputed into \{example\_review\} and \{example\_label\} and actual user review is input to \{user\_review\}. The model generates one token following the prompt, corresponding to the underbar at the end of the prompt.

\begin{figure}[htb]
\begin{center}
\begin{lstlisting}
### Instruction
You are a helpful assistant evaluating the review texts about the restaurant. Please evaluate the review text and assign an integer score ranging from 1 for the most negative comment to 5 for the most positive comment.

### Example
User review: {example_review}
Rating: {example_label}

### Output
User review: {user_review}
Rating: _
\end{lstlisting}
\caption{Basic Prompt}\label{fig:prompt}
\end{center}
\end{figure}

\subsection{One-shot Strategies}
\subsubsection{Centroid-based Representative Examples}
For CREs (proposed), we select representative samples based on the combination of SentenceBERT (SBERT) \cite{sbert} embeddings\footnote{https://huggingface.co/sentence-transformers/all-MiniLM-L6-v2} and their clusters.
First, we obtain fixed-length embeddings for all candidate texts using SBERT.
For each sample, mean pooling is applied over all token embeddings to produce a single 384-dimensional vector representation.

Second, we apply K-Means clustering to the set of embedding vectors, setting the number of clusters $K$ equal to the number of one-shot examples to be selected (e.g., $K=5$ in our experiments).
For each cluster, the sample whose embedding is closest to the cluster centroid (using Euclidean distance) is chosen as a representative.
The final set of $K$ Centroid-based Representative Examples (CREs) thus consists of the centroid-nearest samples from each cluster.
These CREs are then used as one-shot examples in the prompts for the LLM ensemble experiments.

\subsubsection{Randomly-Selected Examples (baseline)}
For RSEs (baseline), we randomly sample $K$ examples from the training pool for each experiment, using a fixed random seed for reproducibility. No clustering or embedding-based selection is applied; the examples are simply drawn at random as a baseline for comparison with the proposed CREs.

\subsection{Aggregation}
To consolidate multiple predictions from base models, we employ median aggregation.
While other aggregation methods, such as majority voting, are also possible, prior work \cite{niimi_nldb} has highlighted several issues with these approaches.
Specifically, in situations where the votes are split among classes, majority voting can lead to ambiguous or unstable results. Moreover, when applied to tasks such as sentiment classification, which are inherently ordinal, majority voting treats the problem as a nominal classification, disregarding the ordinal relationships between labels.
In contrast, median aggregation is more appropriate for ordinal outcomes, as it directly incorporates the order information and is less susceptible to the influence of outlier predictions.

\section{Experimental Study}
\subsection{Dataset and Settings}
This study employs Yelp Open Dataset \cite{yelp} which is publicly available dataset for academic research. The dataset contains user ratings (stars: 1--5) and reviews for a wide range of businesses. We focus on the venues tagged with Restaurants and randomly split the data into two sets: data pool for one-shot examples (sample size: 18,000) and test set (sample size: 1,000). We first extract five examples from data pool and establish five base models. Examples for each one-shot strategy (CREs and RSEs) are chosen from data pool. The summary statistics for the test set is shown in Table \ref{tab:stats} (The rating and characters are shown in {\it mean} $\pm$ {\it std}). 

\begin{table}[htb]
\begin{center}
      \caption{Summary Statistics of the Test Set}\label{tab:stats}
\begin{tabular}{
@{}
 wc{0.4cm}
wc{0.7cm}wc{0.8cm}
wc{1.8cm}wc{2.5cm}
@{}
}
\toprule
n & Users & Venues &  Rating & Characters \\
\midrule
$1000$ & $993$ & $503$   & $3.979 \pm 1.325$ & $480.716\pm 455.688$ \\
\bottomrule
\end{tabular}
\end{center}
\end{table} 

\subsection{Model Evaluation}
This study evaluates the performance of the proposed model using three evaluation metrics: accuracy (Acc.), macro-F1 (F1), and root mean square error (RMSE). Accuracy and macro-F1 are standard classification metrics, whereas RMSE quantifies the magnitude of prediction errors between predicted and true labels. To assess statistical significance, we employ McNemar's test and the Wilcoxon signed-rank test to verify the performance improvements achieved by the proposed models. 

\subsection{Selection of Examples}
\begin{figure*}[tbh]
   \centering
   \includegraphics[width=0.95\linewidth]{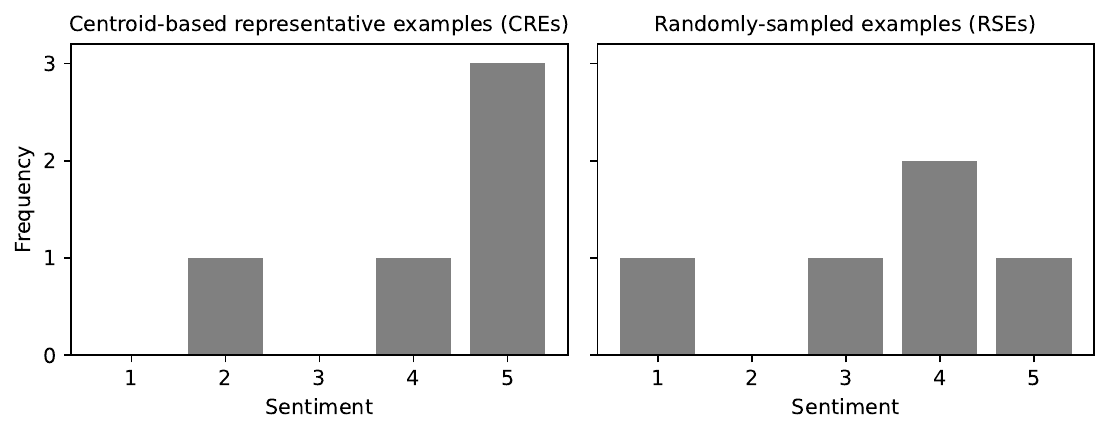}
   \caption{Rating distributions for examples in each group. While the RSEs are relatively balanced across the five strata, the CREs are notably skewed toward higher ratings.}\label{fig:examples}
\end{figure*}

The distributions of ratings depending on sampling strategy is shown in Fig. \ref{fig:examples}. Both strategy basically reflect the overall distributional bias in the original dataset; positive ratings (4--5 stars) are more prevalent than negative ones (1--2 stars). 

For CREs, the positive samples generally emphasized food quality, variety, and service, frequently expressing enthusiasm and intent to revisit. The negative example cited specific issues with service, such as delayed orders and inattentiveness, despite acknowledging food quality. This composition illustrates that, while clustering captures semantic diversity, the inherent skew of the dataset can result in a concentration of positive examples. However, each representative example provides distinct context, such as menu diversity, staff attentiveness, and overall dining experience\footnote{In accordance with Yelp’s data policy, only summary descriptions of the content and not the raw texts are presented in this paper.}.

On the other hand, in the case of RSEs, the examples capture various aspects of restaurant experiences, including delivery delays, issues with menu availability and customer service, food quality and personal taste preferences, as well as overall satisfaction with food and atmosphere. This comparison demonstrates that the CRE-based strategy may increase semantic diversity but label-skewed examples due to dataset bias, while random sampling achieves a more balanced sentiment distribution at the potential cost of context relevance.

\subsection{Results and Discussions}
Table \ref{tab:results} reports the performance of the five individual models (M1--M5) and their median ensemble (Ens) under two one-shot strategies (randomly sampled examples, RSEs; centroid-based representative examples, CREs) and two sampling temperatures $T \in \{0.8, 1.5\}$. 

\renewcommand{\colwid}{1.3cm}
\begin{table*}
\begin{center}
   \caption{Effect of Ensemble Inference. The bold type indicates that the ensemble model outperforms the individual base models and cell shading represents the highest performance. }\label{tab:results}
    \begin{tabular}{
    %@{}
    wc{0.8 cm}wl{1.6 cm}
    wc{\colwid}wc{\colwid}wc{\colwid}wc{\colwid}wc{\colwid}wc{\colwid}wc{\colwid}
    %@{}
    }
\toprule 
\multicolumn{2}{l}{\bf RSEs (baseline)} &M1&M2&M3&M4&M5&Avg.&Ens\\
\midrule
\multirow{3}{*}{0.8} 
&$Acc.$& 0.732 & 0.735 & 0.684 & 0.692 & 0.691 &0.707& 0.735\\
&$F_1$&        0.591 & 0.577 & 0.495 & 0.492 & 0.518 & 0.535 & \bf0.596 \\
&RMSE & 0.559 & 0.578 & 0.670 & 0.682 & 0.658 &0.629& 0.584 \\
\midrule
\multirow{3}{*}{1.5} 
&$Acc.$& 0.728 & 0.715 & 0.667 & 0.501 & 0.699 &0.662& 0.718 \\
&$F_1$  & 0.550 & 0.433 & 0.438 & 0.342 & 0.467 &0.446& \bf0.591\\
&RMSE & 0.628 & 1.114 & 0.718 & 1.274 & 0.667 &0.880& \bf0.572 \\
\midrule \midrule
\multicolumn{2}{l}{\bf CREs (proposed)} &M1&M2&M3&M4&M5&Avg.&Ens\\
\midrule
\multirow{3}{*}{0.8} 
&$Acc.$&  0.735 & 0.695 & 0.689 & 0.728 & 0.687 &0.707& 0.714 \\
&$F_1$&        0.592 & 0.567 & 0.497 & 0.546 & 0.482 &0.537& 0.557\\
&RMSE & 0.568 & 0.598 & 0.642 & 0.630 & 0.636 &0.615& 0.593\\
\midrule
\multirow{3}{*}{1.5} 
&$Acc.$  & 0.693 & 0.756 & 0.661 & 0.275 & 0.665 &0.610& \graycell\bf0.762 \\
&$F_1$  & 0.619 & 0.623 & 0.533 & 0.193 & 0.437 &0.485& \graycell\bf0.636 \\
&RMSE &0.574 & 0.539 & 0.720 & 1.950 & 0.651 &0.887& \graycell\bf0.512\\
\bottomrule
    \end{tabular}
\end{center}
\end{table*}

\newpage
\subsubsection{Effect of Sampling Temperature}
We first examine the effect of sampling temperature (RQ1) by comparing the models within the sampling approaches.
\begin{itemize}
\item[] \textbf{RSEs (baseline).} Raising the temperature from $0.8$ to $1.5$ changes F1 only from 0.596 to 0.591 ($-0.8\,\%$) and RMSE from 0.584 to 0.572 ($-2.1\,\%$).  In other words, random examples provide little additional diversity even at the higher temperature.
\item[] \textbf{CREs (proposed).} In contrast, the same temperature shift improves F1 from 0.557 to 0.636 ($+14.2\,\%$) and reduces RMSE from 0.593 to 0.512 ($-13.7\,\%$). The improvement was statistically significant according to both McNemar’s test on paired correctness ($p<0.05$, $\chi^2=5.84$) and Wilcoxon signed-rank test on absolute errors ($p<0.01$, $W=31534$), confirming that representative examples benefit from larger output variance.
\end{itemize}

These results present the concrete answer to RQ1 (The effect of sampling temperature) that the temperature controls the model diversity: low $T$ leads to the deterministic outputs and ineffective ensemble while high $T$ to diverse outputs and ensemble improvements.

\newpage
\subsubsection{Effect of One-shot Strategy}
We second examine the effect of one-shot strategy by comparing the models within the sampling temperatures (RQ2: ensemble representativeness)
\begin{itemize}
\item[] \textbf{Low temperature.} First, both strategies are statistically indistinguishable: RSE-Ens does not have a significant improvement from the single best model and CRE-Ens is even below RSE-Ens. In the lower temperature setting, predictions become nearly deterministic and the median offers no advantage. 
\item[] \textbf{High temperature.} With the higher temperature setting, sufficient diversity is provided to each model. As the results, CRE-Ens clearly outperforms RSE-Ens: F1 improves from 0.591 to 0.636 ($+7.61\,\%$) and RMSE reduces from 0.572 to 0.512 ($-10.49\,\%$). 
We confirm that this improvement is also statistically significant according to both McNemar's test on paired correctness ($p<0.05$, $\chi^2=5.193$) and Wilcoxon signed-rank test on absolute errors ($p<0.01$, $W=28408$).
\end{itemize}

These results also show the clear answer to RQ2 (representativeness) that the representativeness of the example has a significant effect on ensemble performance with the sufficient diversity.

\subsubsection{Overall}
Across both metrics, the CRE-Ens at $T=1.5$ achieves the highest F1 (0.636) and the lowest RMSE (0.512), surpassing all other ensemble and individual models, indicating the clear answer to RQ3 (optimal configuration) that CRE-approach with high temperature setting realizes the optimum balance between representativeness and diversity.

Table \ref{tab:details} presents the detailed results of per-class and overall F1 scores for each base model ($M_1$--$M_5$) and the ensemble (ens) for CREs (proposed) with high temperature setting. Due to the imbalanced class distribution, F1 for individual classes can remain $0$ or reach $1$ in some base models. This phenomenon is especially pronounced in multiclass sentiment classification tasks, where the precision and recall for certain classes may fluctuate significantly depending on the model’s output distribution.
While the ensemble achieves the highest overall F1, it also smooths out the class-specific extremes observed in individual models. These results highlight the inherent difficulty of achieving balanced performance across all classes in imbalanced, real-world datasets.

These results provide the clear answer to RQ3 (Optimal combination) that the combination of centroid-based representative examples (CREs) and high sampling temperature ($T=1.5$) yields the best overall ensemble performance (F1~$ = 0.636$, RMSE~$ = 0.512$), outperforming all other configurations.

\renewcommand{\colwid}{0.8cm}
\begin{table}[htb]
   \centering
      \caption{F1 by the Labels for CREs ($T=1.5$)}\label{tab:details}
\begin{tabular}{
@{}
wc{0.7cm}
wc{\colwid}wc{\colwid}wc{\colwid}wc{\colwid}wc{\colwid}wc{\colwid}
@{}
}
\toprule
Labels & 1 & 2 & 3 & 4 & 5 & All \\
\midrule
$M_1$ & 0.320 & 0.201 & 0.125 & 0.232 & 0.398 & 0.619 \\
$M_2$ & 0.488 & 0.183 & 0.116 & 0.195 & 0.317 & 0.623 \\
$M_3$ & 0.226 & 0.138 & 0.237 & 0.003 & \bf1. & 0.533 \\
$M_4$ & 0.120 & \bf1. & 0. & 0.008 & \bf0.286 & 0.193 \\
$M_5$ & \bf0.493 & 0.086 & 0.008 & 0.073 & 0.498 & 0.437 \\
Ens & 0.480 & 0.201 & 0.116 & 0.199 & 0.475 & \bf0.636 \\
\bottomrule
\end{tabular}
\end{table} 

\subsubsection{Comparison with 5-shot Model}
To validate RQ4 (Comparison with 5-shot model), we construct two five-shot models with different temperature settings $T \in \{0.8, 1.5\}$.  Table \ref{tab:fiveshot} compares the performance between five-shot single models and the best ensemble model (Ens: CREs with $T = 1.5$). While five-shot prompting offers a slight improvement over the average of one-shot base models, it remains substantially inferior to the ensemble approach. 

\renewcommand{\colwid}{1.4cm}
\begin{table}[h]
   \centering
      \caption{Comparison with 5-shot Models for CREs. Lift indicates the performance increase from the best 5-shot model.}\label{tab:fiveshot}
\begin{tabular}{
@{}
wl{0.7cm}
wc{\colwid}wc{\colwid}wc{\colwid}wc{\colwid}
@{}
}
\toprule
& \ccol{2}{5-shot (CREs)} & \multirow{2}{*}{CRE-Ens} & \multirow{2}{*}{Lift} \\
\cmidrule(lr){2-3}
& $T=0.8$ & $T=1.5$ & &\\
\midrule
 Acc.   & 0.692 & 0.642 & \bf0.762 & $+10.1\%$  \\
 F1   &  0.525 & 0.496 & \bf0.636 & $+21.1\%$ \\
 RMSE & 0.674 & 0.804 &  \bf0.512 & $-24.0\%$ \\
\bottomrule
\end{tabular}
\end{table} 

A McNemar’s test on paired correctness confirmed that the differences were statistically significant for both temperature settings ($p < 0.001$, $\chi^2 = 15.87$ for $T = 0.8$ and $\chi^2 = 43.17$ for $T = 1.5$). Wilcoxon signed-rank test on absolute errors also confirmed the significant difference ($p<0.001$, $W=14245$ for $T=0.8$ and $W=16457$ for $T=1.5$).
This underscores that ensemble aggregation is crucial for harnessing the diversity introduced by high temperature, and simply increasing the number of prompt examples is not sufficient to achieve robust performance.

\subsubsection{Self-consistency}
Finally, we validate the self-consistency of our best ensemble model. Table \ref{tab:consistency} summarizes the agreement among the five base models. $n_{\text{unique}}$ denotes the number of unique predicted labels produced by the five models for each sample (i.e., $n_{\text{unique}} = 1$ means all models predict the same label, while $n_{\text{unique}} \geq 2$ indicates disagreement). The case where all models made different predictions ($n_{\text{unique}} = 5$) did not occur. We further stratify the samples by $n_{\text{unique}}$ and compute the F1 score for each stratum.

\renewcommand{\colwid}{0.9cm}
\begin{table}[h]
   \centering
      \caption{Self-consistency and performance for CREs ($T=1.5$)}\label{tab:consistency}
\begin{tabular}{
@{}
wl{1.3cm}
wc{\colwid}wc{\colwid}wc{\colwid}wc{\colwid}wc{\colwid}
@{}
}
\toprule
$n_{\text{unique}}$ & 1 & 2 & 3 & 4 & 5 \\
\midrule
Samples & 225 & 328 & 427 & 20 & 0 \\
F1 Score & 0.938 & 0.756 & 0.684 & 0.550 & - \\
\bottomrule
\end{tabular}
\end{table} 

The model performance improves for samples where all base models agree ($n_{\text{unique}} = 1$), indicating that high consistency among the base models increases the model performance. Even in the low-consistency cases, as shown in the effect of ensemble, ensemble method can significantly reduce the prediction error with median and outperforms individual models.

These results clearly answer RQ5 (Effect of consistency) that, in samples with high self-consistency, ensemble methods enable high confidence in correct answers, while in samples with low self-consistency, ensemble methods correct errors. Thus, in both cases, ensemble method plays an important role to ensure the performance. As pointed out in prior study \cite{llm_selfconsistency}, self-consistency is the important factor to improve the prediction performance in LLMs. 

\section{Conclusion}
\subsection{Summary}
In this study, we conducted a sentiment analysis using an ensemble approach of multiple LLMs, systematically varying both the one-shot example selection strategy and the sampling temperature parameter to identify the optimal ensemble configuration. Through addressing five RQs, this study has three key findings:
i.) the proposed centroid-based representative examples (CREs) with a high temperature setting demonstrated substantial improvements over several baselines. 
ii.) our best ensemble model significantly outperformed the 5-shot model whose examples are selected with the proposed approach. 
iii.) the proposed ensemble approach is particularly beneficial for samples where individual models disagree (low consistency), while also reinforcing confidence in cases of high model agreement (high consistency). 
These findings highlight the practical importance of both representative example selection and controlled diversity in maximizing the benefit of LLM ensembles.

In summary, our results demonstrate that: (1) higher temperature increases diversity and, together with effective example selection, enhances ensemble performance; (2) centroid-based representative examples yield superior results over random examples; (3) the optimal ensemble configuration combines CREs and high temperature; and (4) ensemble aggregation is most beneficial for samples with low self-consistency, but also boosts confidence in cases of high agreement.

Finally, the proposed ensemble achieved a $+2.1\%$ improvement in F1 over the best individual base model. While obtaining such a strong base model typically requires extensive exploration of exemplar combinations, our method automatically selects representative examples through clustering, eliminating the need for manual search or domain-specific tuning. This demonstrates both the efficiency and practicality of the proposed approach. 

\subsection{Implications}
This study has several implications both for academic and practical aspects.

\subsubsection{Theoretical Implications}
Our findings show that simply inserting a representative one-shot example is not, by itself, sufficient to raise ensemble performance. Rather, ensemble accuracy is mediated primarily by the temperature parameter, and the magnitude of that effect depends strongly on the example-selection strategy. Notably, even a single model reached an F1 of 0.619, exceeding all other ensemble models. Random sampling disperses the decision boundary, whereas centroid examples keep the base models in the same direction, letting the median discard outliers.

Although the ensemble built from CREs with $T=1.5$ achieved the highest overall accuracy, the constituent models were not individually strongest at that same temperature. This observation is consistent with the accuracy--diversity trade-off \cite{ensemble_tradeoff}, showing that increasing diversity often lowers individual accuracy, yet may still improve the ensemble as a whole. Our results confirm that ensemble performance does not hinge on the superiority of each component model: the best ensemble still emerged even when it incorporated models with individual macro-F1~$=0.193$.

Regarding the self-consistency among the individual decisions, the observed relationship between model consistency and ensemble performance has important implications for the design of LLM-based classification systems. First, our results confirm that self-consistency is a reliable indicator of prediction confidence: samples with unanimous agreement among base models are nearly always classified correctly, which aligns with prior findings on the value of self-consistency in LLM inference \cite{llm_selfconsistency}. Second, the main benefit of ensemble aggregation lies in its ability to recover correct predictions in difficult cases where base models disagree, thereby improving overall robustness.
These findings suggest that future systems could leverage self-consistency as a confidence metric, enabling more targeted use of ensemble inference or adaptive post-processing for ambiguous samples. Our empirical analysis thus bridges theoretical discussions on LLM consistency with practical ensemble design for real-world sentiment analysis.

Importantly, our findings indicate that example selection based on semantic clustering does not necessarily align with the original label distribution; for instance, the selected centroid-based examples skewed towards high sentiment ratings (Fig. \ref{fig:examples}). Nevertheless, this approach captures a broad range of semantic contexts and yields sufficient diversity to enable effective ensemble inference. This suggests that semantic diversity among prompt examples, rather than strict label diversity, is a key factor for maximizing the robustness and performance of LLM ensembles.

\subsubsection{Practical Implications}
By clustering SBERT embeddings and automatically selecting the cluster centroids as one-shot examples, the proposed approach enables practitioners to construct effective LLM ensembles without domain-specific prompt engineering or extensive manual intervention. Since the proposed sampling strategy (CRE) only affects the prompting process, while the temperature setting only modifies the sampling distribution during generation, the overall processing time remains almost identical across configurations.  Therefore, companies can easily improve performance without implementing complex inference methods or incurring additional computational costs. This lowers the barrier to deploying robust sentiment analysis models in industrial applications, where labeled data may be limited and human resources for model tuning are scarce.

Furthermore, the use of self-consistency as an indicator of prediction confidence provides a practical means to automatically flag ambiguous or difficult cases for further review or targeted intervention, which is valuable in applications such as customer feedback analysis, automated moderation, or decision support.

Finally, the demonstrated effectiveness of the ensemble with centroid-based example selection and temperature-driven diversity suggests that similar strategies could be readily adapted for other text classification or ordinal prediction tasks, further extending the practical utility of the method beyond sentiment analysis.

\subsection{Limitations}
As for the limitations, our proposed methodology is validated with the single dataset of 1,000 samples, with the case of five base models for the ensemble; the effect and scalability of increasing the number of base models require further cross-domain validation, such as finance and healthcare, using other pretrained models, such as Qwen \cite{qwen3} and Mistral \cite{mistral}.

Also, this study does not examine the relationship between the self-consistency and confidence or reliability of the inference in detail. Particularly, leveraging self-consistency as a confidence indicator enables dynamic inference strategies. Further theoretical analysis on the accuracy--diversity trade-off is worth investigating.

\section*{Acknowledgments}
Both the dataset and model were managed and used in an appropriate environments that comply with the terms of use. We do not collect additional information that could lead to the identification of individuals. 

This study is supported by JSPS KAKENHI (Grant Number: 24K16472).

\bibliographystyle{unsrt}
\bibliography{representshot}

\end{document}